\documentclass[review,number,sort&compress,12pt]{elsarticle}%

\usepackage{natbib}
\usepackage{geometry}
\usepackage{graphicx}
\usepackage{subfigure}
\usepackage{scrtime}
\usepackage{epstopdf}
\usepackage{enumerate}
\usepackage{amsmath}
\usepackage{siunitx}
\usepackage{amssymb}
\usepackage{upgreek}
\usepackage{appendix}
\usepackage[colorlinks, linkcolor=black, citecolor=black, anchorcolor=black]{hyperref}
\hypersetup{hidelinks}
\usepackage{algorithm}% http://ctan.org/pkg/algorithms
\usepackage{algorithmicx}
\usepackage{algpseudocode}% http://ctan.org/pkg/algorithmicx

\usepackage{multirow}
\usepackage{lineno}
\usepackage{enumerate}
\usepackage[section]{placeins}

\usepackage{booktabs}
\usepackage{threeparttable}

\usepackage{booktabs}
\usepackage[dvipsnames]{xcolor}
\usepackage{tcolorbox}
\usepackage{soul}
\usepackage{makecell}
\usepackage{mathtools}
\def\romannumer#1{\uppercase\expandafter{\romannumeral#1}}
\begin{document}
% \modulolinenumbers[1]
% \linenumbers

\begin{frontmatter}
%Title of paper
\title{Physics-informed machine learning of redox flow battery based on a two-dimensional unit cell model}

\author[1]{Wenqian Chen}
%\ead{wenqian.chen@pnnl.gov}

\author[1]{Yucheng Fu}
%\ead{yucheng.fu@pnnl.gov}

\author[1]{Panos Stinis \corref{cor1}}
\ead{panos.stinis@pnnl.gov}

\cortext[cor1]{Corresponding author}

\address[1]{Advanced Computing, Mathematics and Data Division \\
	Pacific Northwest National Laboratory \\ Richland, WA 99354, USA}

\begin{abstract}
In this paper, we present a physics-informed neural network (PINN) approach for predicting the performance of an all-vanadium redox flow battery, with its physics constraints enforced by a two-dimensional (2D) mathematical model. The 2D model, which includes 6 governing equations and 24 boundary conditions, provides a detailed representation of the electrochemical reactions, mass transport and hydrodynamics occurring inside the redox flow battery. To solve the 2D model with the PINN approach,  a composite neural network is employed to approximate species concentration and potentials; the input and output are normalized according to prior knowledge of the battery system; the governing equations and boundary conditions are first scaled to an order of magnitude around 1, and then further balanced with a self-weighting method. Our numerical results show that the PINN is able to predict cell voltage correctly, but the prediction of potentials shows a constant-like shift. To fix the shift, the PINN is enhanced by further constrains derived from the current collector boundary. Finally, we show that the enhanced PINN can be even further improved if a small number of labeled data is available. 
	
\end{abstract}


\end{frontmatter}

\section{Introduction}
Redox flow batteries (RFBs) have attracted increasing attention in recent years due to their promising applications in large-scale energy storage systems \cite{soloveichik2015flow, noack2015chemistry, weber2011redox}. RFBs are characterized by their decoupled energy and power capacities, long cycle life, and high efficiency, which make them ideal for integrating intermittent renewable energy sources such as wind and solar power \cite{shah2008dynamic,leung2017recent}. However, the widespread deployment of RFBs is hindered by several challenges, including limited energy density, complex reaction kinetics, and high system costs \cite{kim20131, skyllas2019performance}. Accurate and efficient prediction models for RFB performance are critical for addressing these challenges and enabling the optimization of battery system design and operation. In this work, we focus on the prediction of the performance for an all-vanadium redox flow battery (VRFB) \cite{sum1985study, skyllas1987efficient}, which is
one of the most popular and mature RFB technologies \cite{kear2012development}.

In recent years,  numerical models are playing an increasingly important role in the understanding and prediction  of  VRFBs performance. Physics-based VRFB models, frequently used in the field, come in a variety of forms. They range from the  zero-dimensional (0D) models \cite{shah2011dynamic, sharma2015verified, eapen2019low, lee2020open}, through the single-axis dynamics of one-dimensional (1D) models \cite{vynnycky2011analysis, chen2014enhancement}, the plane-play of two-dimensional (2D) models \cite{shah2008dynamic, sharma2014quasi, al2009non, you2009simple, shah2010dynamic, knehr2012transient, choi2020multiple}, to the comprehensive spatial interactions of three-dimensional (3D) models \cite{ma2011three, xu2013numerical, fu2023three, zheng2014three, yin2014coupled, oh2015three, yin2015numerical, messaggi2018analysis}. The type of model chosen hinges on the number of spatial dimensions in which species concentrations, current density, flow velocity and potential are set to fluctuate. The choice of a model depends on the balance between accuracy and computational cost. High-dimensional models deliver greater accuracy but demand more computational resources, while simpler models are less resource-intensive but may compromise on accuracy. Conventional numerical models employ techniques like finite difference, finite element, or finite volume methods to discretize the computational domain of the battery cell. As models and simulations advance to higher dimensions, the intricacy of discretization increases exponentially, presenting additional computational obstacles and complexity.

Machine learning (ML) techniques \cite{goodfellow2016deep} have emerged as a powerful alternative to traditional modeling approaches, providing fast and accurate predictions for various energy storage systems \cite{artrith2019machine, gao2021machine,  chen2020machine}. Among ML techniques,  data-driven neural networks (NNs) have demonstrated promising performance in predicting the behavior of complex systems, even when an in-depth understanding of the underlying physical processes is lacking.  For applications in RFBs, neural networks are usually employed to build surrogates, and have been applied to the optimization of  electrode structures \cite{wan2021coupled}, the optimization of performance and cost \cite{li2020cost}, and  the investigation about the influence of pore-scale electrode structure on device-scale electrochemical reaction uniformity \cite{bao2020machine}.
However, data-driven neural networks often rely solely on large amounts of experimental/simulation data, which can be challenging to obtain for a wide range of operating conditions. Additionally, their lack of interpretability can make it difficult to gain insights into the underlying physical processes or to identify the important parameters that governs the RFB performance. 

Physics-informed machine learning  or physics-informed neural networks (PINNs) in particular represents a significant advancement in the field, as they combine the power of deep learning with the incorporation of physical constraints. 
Recent studies have underscored the benefits of physics-informed machine learning (ML), particularly its superiority over purely data-driven neural networks, as it incorporates important physical laws. 
{\color{black}
Most of the recent studies have focused on lithium-ion batteries, where physics-informed machine learning has been successfully applied to thermal process prediction\cite{pang2023physics,dakshinamoorthy2023estimating, deng2023physics, cho2022physics}, health prognosis\cite{wen2023fusing,sun2023adaptive}, and performance prediction \cite{huang2023minn, wang2022physics}. 
As for VRFBs, a PINN approach was developed in \cite{he2022physics},} which incorporated constraints from a 0D model, to infer the parameters of a VRFB model as well as its voltage curves. The outcomes suggest that the inclusion of physical constraints enables the PINN to deliver important insights into the physical processes of VRFBs.
\cite{he2022enhanced} continued their work and introduced an improved version of the PINN, which significantly improved discharge voltage curve tail region accuracy. Those study demonstrated that the PINN's integration of physical laws allows for more precise predictions and make the extrapolation of those neural networks more robust.
In another work\cite{howard2022physics},  a multi-fidelity machine learning approach based on Gaussian processes is developed for cell voltage prediction of VRFBs,  where a 0D model is employed as low-fidelity constraints.  The findings revealed that only a small amount of data is necessary for accurate predictions, suggesting a practical potential for scenarios where experimental data may be scarce or costly to obtain.

{\color{black}The primary limitation of physics-informed machine learning is the necessity for a well-defined physics-based model of the problem in focus. Analogous to conventional numerical models, the accuracy and computational cost of physics-informed machine learning increase with the fidelity of the employed physics-based model.} Although significant advancements have been made in the field of physics-informed machine learning, its application to VRFBs has primarily been confined to 0D models, due to their superior computational efficiency. However, the accuracy and adaptability of these 0D models are compromised due to the limitations imposed by the lumped-parameter assumption and simplification of the battery geometry. 
In this study, we utilize a PINN to tackle the complexity inherent in a 2D physics-based device-scale VRFB model. This 2D model effectively encapsulates the intricate dynamics between electrochemical reactions and mass transport within the VRFB system. 
Within this approach, the two-dimensional model is integrated as physical constraints, which are subsequently resolved via the procedure of neural network training. Importantly, this process circumvents the typical requirement for spatial discretization characteristic of conventional numerical methods. Additionally, this methodology provides the advantage of incorporating further prior knowledge, such as the bounds of solution, additional physical laws, or labeled data.
Once the neural network has been trained, this approach provides a robust and efficient means for predicting the performance of VRFBs. This  application of PINNs  paves the way for a more comprehensive understanding and optimization of VRFBs by leveraging the promise of deep learning. {\color{black}In addition, prominent energy storage technologies like lithium-ion, lithium-sulfur, and solid-state batteries have in-depth, physics-based numerical models \cite{bates2015modeling, santhanagopalan2006review, abadi2016tensorflow, jokar2016review, fu2023understanding, kumaresan2008mathematical, danilov2010modeling} developed by various researchers to illuminate their complex systems and dynamics. Following the same procedures proposed in this work, the PINN method introduced in this research can be adapted to these energy storage systems to predict their performance.}

\section{Methods}
\subsection{Mathematical model }
\label{sec_2Dmodel}
\begin{figure}[H]
	\centering	
	\includegraphics[width=14.4cm]{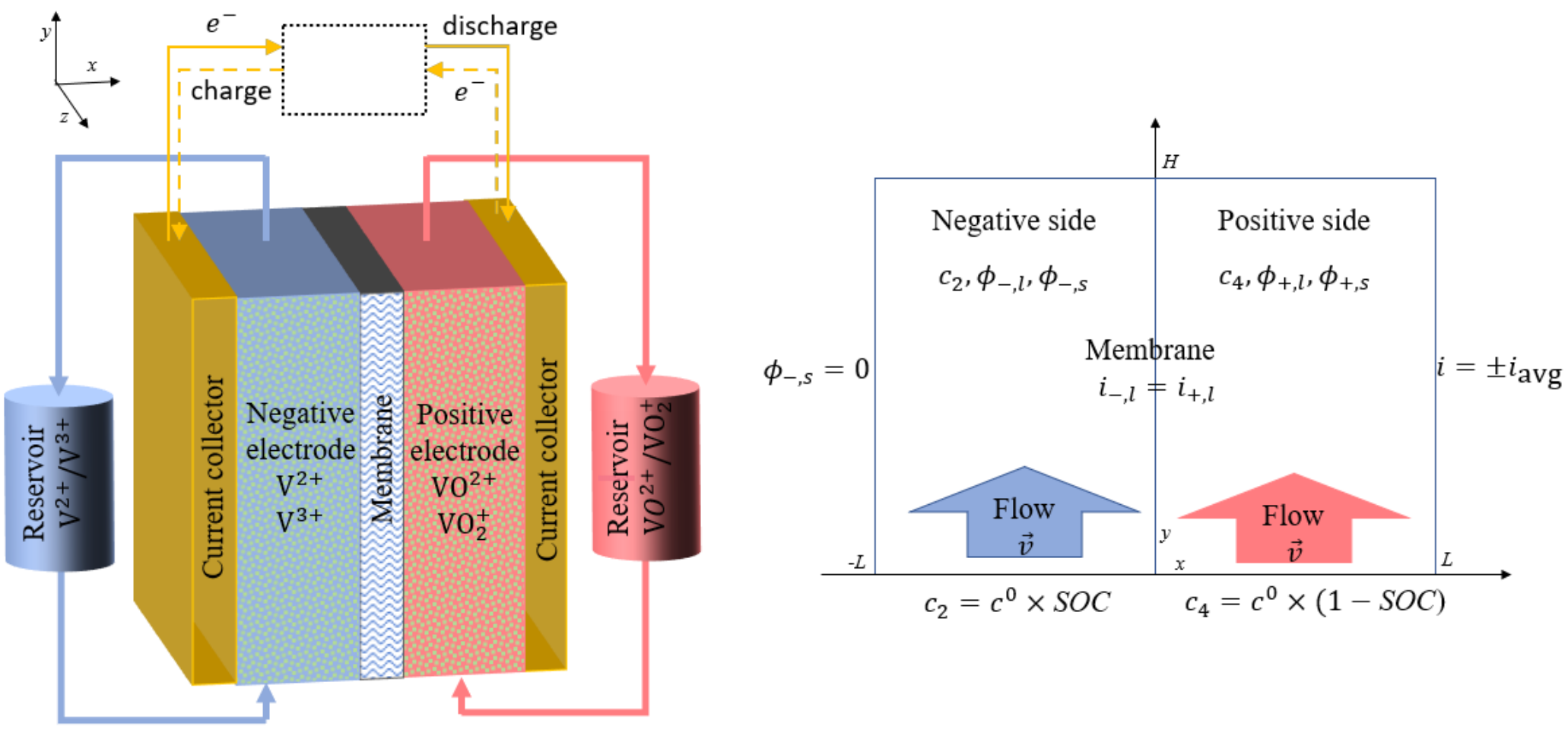}
	\caption{Schematic (left) and mathematical  model (right) of a single-cell all-vanadium redox flow battery. Here $c^0$ is the total vanadium concentration of the positive/negative side.}
	\label{fig_model}	
\end{figure}

The schematic of a single-cell all-vanadium redox flow battery is shown in Fig. \ref{fig_model}. The cell contains porous electrode with electrolyte for both the negative and positive sides. The two electrodes are separated by a proton exchange membrane (PEM) that allows protons to pass through while preventing the vanadium species from mixing. During charging, $\text{V}^{3+}$ in the negative electrolyte is
converted to $\text{V}^{2+}$ and {\color{black} $\text{VO}^{2+}$ in the positive electrolyte is converted to $\text{VO}_2^{+}$.}
This process involves the transfer of  protons across the PEM, converting electrical energy into chemical energy stored in the vanadium species. During discharging, the flow direction of electrons and protons is reversed, generating electrical energy.

In order to build a manageable mathematical model, several assumptions are adopted for simplification. Here, we only consider the components of membrane, negative/positive electrodes and electrolytes. As shown in the right part of Fig. \ref{fig_model}, the positive and negative sides are assumed to be 2D rectangular domains (height $H$ and length $L$)  with negligible variations in the width direction. The two-dimensional domain approximation only sacrifices  a small part of the accuracy when the electrolyte flow rate is reasonably high, as pointed out in \cite{shah2008dynamic}.  The membrane is simplified as a zero-thickness interface connecting the negative and positive electrodes.  The cycling of the electrolyte species and the reservoir are modeled by specifying the inlet species concentration according to the state of charge (SOC) and the inlet electrolyte velocity. During the charging and discharging stages, the current density remains constant as $-i_{avg}$ and $i_{avg}$, respectively. The term $i_{avg}$ is the average current density over the current collectors, which is defined as
\begin{equation}
	i_{avg} = \frac{I}{HW}
\end{equation}
where $I$ is the total current and $W$ is the cell width. The fluid flow of the negative and positive electrolytes is assumed to be uniform with a constant flow velocity $\vec{v}$.
% \vspace{-0.5cm}

\subsubsection{Equations for the electrodes}
The kinetics for VFRB are highly complex, including not only the electrochemical reactions of vanadium species but also several potential side effects in extreme conditions (such as hydrogen evolution and oxygen evolution). We only consider the main reaction kinetics as follows:
\begin{align}
	&\text{Negative side:}
	\quad
	 &\text{V}^{3+} + e^{-} &\xrightleftharpoons[\text{Discharge}]{\text{Charge}} \text{V}^{2+}
	 \label{eq_reaction_neg} 
	\\
	&\text{Positive side:}
	\quad
	&\text{VO}^{2+} + \text{H}_2\text{O} - e^{-} &\xrightleftharpoons[\text{Discharge}]{\text{Charge}}
	\text{VO}_2^+ + 2\text{H}^+
	\label{eq_reaction_pos}
\end{align} 

The transport of chemical species inside the negative and positive electrode pore channels is described by the Nernst-Planck equation
\begin{equation}	
	\label{eq_NP}
	\frac{\partial}{\partial t}(\epsilon c_j) + \nabla \cdot \vec{N}_j = -S_j
	,
\end{equation}
where $c_j$ is the concentration of species $j$, $\epsilon$ is porosity, and $S_j$ is the source term for species $j$ (defined in Table \ref{table_sources_NP}). Note that we use the subscripts $j=$ 2, 3, 4 and 5 to denote species $\text{V}^{2+}, \text{V}^{3+}, \text{VO}^{2+}$,  and $\text{VO}_2^+$, respectively. The concentration flux $\vec{N}_i$ is defined as
\begin{equation}	
	\label{eq_NP_flux}
	\vec{N}_j = \vec{v}c_j-D_j \nabla c_j-\frac{z_j D_j c_j}{RT}F\nabla \phi_l
	,
\end{equation}
which comprises of hydrodynamic convection, diffusion and electrostatic force. $D_j$ and $z_j$ are the diffusion coefficient and valence of the species $j,$ $\phi_l$ is the electrolyte potential, $R$ is the universal gas constant, $T$ is the temperature, and $F$ is the Faraday constant. The concentration of species $\{\text{H}^+, \text{H}_2\text{O}, \text{HSO}_4^{-}\}$ is either defined  as a constant or calculated as a functions of $SOC$, as shown in Table \ref{table_params}. The concentration of $\text{SO}_4^{2-}$ is calculated with  the condition of electroneutrality, namely
\begin{align}
	\sum_{i\in \mathcal{C}_{\pm} }{z_ic_i} =0
\end{align}
where $\mathcal{C}_{-}=\{\text{V}^{2+}, \text{V}^{3+}, \text{H}^+, \text{H}_2\text{O}, \text{HSO}_4^{-}, \text{SO}_4^{2-}\}$ is the species collection for the negative electrolyte and  $\mathcal{C}_{+}=\{\text{VO}^{2+}, \text{VO}_2^{+}, \text{H}^+, \text{H}_2\text{O}, \text{HSO}_4^{-}, \text{SO}_4^{2-}\}$ is the the species collection for the positive electrolyte.  The parameter values used in the 2D mathematical model are also given in Table \ref{table_params}.

Based on the scaling analysis in \cite{sharma2014quasi},  the temporal derivative term in Eq. \eqref{eq_NP} can be neglected. This results in a quasi-steady state, with the time-dependence being accounted for solely through the inlet boundary condition.
As pointed out in \cite{you2009simple, qiu20123}, the contribution of the electrostatic force to the concentration distribution of the species is negligible. Thus we remove the electrostatic force term from Eq. \eqref{eq_NP_flux}. Using the reaction for the negative electrode in Eq. \eqref{eq_reaction_neg}, it is deduced that the decrease of V$^{2+}$ concentration is equal to the increase of V$^{3+}$ concentration, and thus the sum of them is a constant. Therefore, we only solve for the V$^{2+}$ concentration, namely $c_2$, for the negative electrode to save on the computational cost. Similarly, we only solve for the VO$^{2+}$ concentration, namely $c_4$, for the positive electrode. In summary, the governing equation for species concentrations to be solved for the negative and positive electrolytes are recast as follows:
\begin{align}	
	&\text{Negative side:}
	\quad
	&\vec{v} \cdot \nabla c_2 -D_2 \nabla^2 c_2 &= -S_2 
	\label{eq_c2}
	\\
	&\text{Positive side:}
	\quad	
	&\vec{v} \cdot \nabla c_4 -D_4 \nabla^2 c_4 &= -S_4
	\label{eq_c4}
\end{align}

\begin{table}[tbp]
	\newcommand{\tabincell}[2]{\begin{tabular}{@{}#1@{}}#2\end{tabular}}
	\centering
	\caption{Source terms defined in Eq. \eqref{eq_NP}.  Here $j_{-,tot}$ and $j_{+,tot}$ are the total current density transferred from solid phase to electrolyte for the negative and positive electrodes, respectively.}
	\begin{tabular}{ccc}
		\toprule
		Electrode & Source term & Value \\
		\midrule
		\multirow{2}{*}{Negative}   & $S_2$ for V$^{2+}$  & $ j_{-,tot}/F$\\
		& $S_3$ for V$^{3+}$ & $- j_{-,tot}/F$\\
		\multirow{2}{*}{Positive}   & $S_4$ for {VO}$^{2+}$ & $ j_{+,tot}/F$\\
		& $S_5$ for {VO}$_2^{+}$  & $- j_{+,tot}/F $\\		
		\bottomrule
	\end{tabular}
	\label{table_sources_NP}
\end{table}

\subsubsection{Electrochemical reactions}
The total current density $j_{\pm, tot}$ in Eq. \eqref{eq_NP} is calculated by the overpotential and species concentration,  which is modeled by the  Butler-Volmer equation
\begin{align}	
	j_{-,tot} &= F a k_{-} c_2^{\alpha_{-,c}} c_3^{\alpha_{-,a}}\left[ \exp\left(   \frac{\alpha_{-,a} F \eta_{-}}{RT} \right)
	-\exp\left( -\frac{\alpha_{-,c} F \eta_{-}}{RT} \right)	
	\right] 
	\label{eq_jtot1}
	\\
	j_{+,tot} &= F a k_+ c_4^{\alpha_{+,c}} c_5^{\alpha_{+,a}}\left[          \exp\left(  \frac{\alpha_{+,a} F \eta_{+}}{RT} \right)
	-\exp\left( -\frac{\alpha_{+,c} F \eta_{+}}{RT} \right)	
	\right]	
	\label{eq_jtot2}
\end{align}
where $a$ is the electrode specific area, $\alpha_{\pm,c}$ is the transfer coefficient and $k_{\pm}$ is the standard reaction rate constant. Note that we use the quantity with subscript ``+/--" to denote the positive/negative cell side, respectively, if not stated otherwise.
The overpotential $\eta _{\pm}$  is denoted by 
\begin{align}
	\eta _\pm &= \phi_{\pm,s} - \phi_{\pm,l} - E_\pm
\end{align}
where $E_{\pm}$ is the open circular voltage (OCV) modeled by the Nernst equation as follows:
\begin{align}
	E _- &=  E_-^0 + \frac{RT}{F}\ln(\frac{c_3}{c_2}) 
	\label{eq_OCV_neg}
	\\
	E _+ &=  E_+^0 + \frac{RT}{F}\ln(\frac{c_5 c_{H^+}^2}{c_4 c_{H_2O}})
\end{align}
The term $E_{\pm}^0$ is the standard OCV for the positive/negative electrode reactions, respectively. 
The terms $\phi_{\pm,s}$ and $\phi_{\pm,l}$ are the electric potential of electrode and electrolyte, respectively. They are determined based on the conservation of charge, namely the charge entering the electrolyte is equal to the charge leaving the solid phase,
\begin{equation}
	\nabla \cdot \vec{i}_{\pm,l} = -	\nabla \cdot \vec{i}_{\pm,s} = j_{\pm, tot}
	\label{eq_current_eq}
\end{equation}
Substituting the definition of the current density, we have the governing equations for the electrolyte and electrode potentials:
\begin{align}
	\nabla \cdot \vec{i}_{\pm,l} &= -\sigma_{\pm,l}^{\text{eff}}\nabla^2 \phi_{\pm,l} =  j_{\pm, tot} 
	\label{eq_phil}
	\\
	\nabla \cdot \vec{i}_{\pm,s} &= -\sigma_{\pm,s}^{\text{eff}} \nabla^2 \phi_{\pm,s} = -j_{\pm,tot} 
	\label{eq_phis}
\end{align}
The terms $\sigma_{\pm,l}^{\text{eff}}$ and $\sigma_{\pm,s}^{\text{eff}} $ are the effective electrolyte and electrode conductivity, which are defined as:
\begin{align}
	\sigma_{\pm,s}^{\text{eff}} & = (1-\epsilon)^{1.5} \times \sigma_s \\
	\sigma_{\pm,l}^{\text{eff}} & = \epsilon^{1.5} \times k_{\pm}^{\text{eff}} 
\end{align}
where $\sigma_s$ is the electrode conductivity. 
$k_{\pm}^{\text{eff}}$ is a function of species concentration and is defined as
\begin{align}
	k_{\pm}^{\text{eff}} &=\frac{F^2}{RT} \sum_{i\in \mathcal{C}_{\pm} }{z_i^2D_ic_i}
\end{align}

\subsubsection{Equations for the membrane}
As shown in the right part of Fig. \ref{fig_model}, the membrane is modeled as a zero-thickness interface. The negative and positive electrodes are coupled through the conservation of current density, namely
\begin{equation}
	 \sigma_{-,l}^{\text{eff}} \frac{\partial \phi_{-,l}}{\partial x}  
	= \sigma_{+,l}^{\text{eff}} \frac{\partial \phi_{+,l}}{\partial x} 
	= \sigma_m	\frac{\phi_{+,l}-\phi_{-,l}}{d_m}
	\label{eq_membrane}
\end{equation}
where $d_m$ is the actual thickness of the membrane and $\sigma_m$ denotes the conductivity of the membrane. 

\subsubsection{Boundary conditions}
The species concentration  at the inlet is defined according to $SOC$, namely
\begin{align}
	c_2=c_2^{in} &= c^0 \times SOC
	\label{eq_inlet_c2} \\
	c_4=c_4^{in} &= c^0 \times (1-SOC) 
	\label{eq_inlet_c4}
\end{align}
where $c^0$ is the total vanadium concentration of the positive/negative side.
The right side of the positive electrode is enforced with a fixed current density, namely
\begin{equation}
	\sigma_{+,s}^{\text{eff}} \frac{\partial \phi_{+,s}}{\partial x}
	= \pm i_{avg} \qquad x=L
	\label{eq_i_avg}
\end{equation}
The electric potential at the negative collector is set as zero, namely
\begin{equation}
	\phi_{-,s}=0 \qquad x=-L
	\label{eq_0_phi_s}
\end{equation}
The electric current density passing through the membrane from the electrode is zero, namely
\begin{equation}
	\sigma_{-,s}^{\text{eff}} \frac{\partial \phi_{-,s}}{\partial x} 
	= \sigma_{+,s}^{\text{eff}} \frac{\partial \phi_{+,s}}{\partial x}
	=  0 \qquad x=0
	\label{eq_bc_membrane}
\end{equation}
All the remaining boundary conditions conform to Neumann type, namely
\begin{align}
	\frac{\partial c_2}{\partial x} = 0 \qquad &x=-L 
	\label{eq_neumann1}
	\\
	\sigma_{-,l}^{\text{eff}}\frac{\partial \phi_{-,l}}{\partial x} =0 \qquad &x=-L
	\label{eq_neumann2}
	\\
	\frac{\partial c_4}{\partial x} = 0 \qquad &x=L 
	\label{eq_neumann3}
	\\
	\sigma_{+,l}^{\text{eff}}\frac{\partial \phi_{+,l}}{\partial x} =0 \qquad &x=L
	\label{eq_neumann4}
	\\		 
	\frac{\partial \phi_{\pm,l}}{\partial y} =\frac{\partial \phi_{\pm,s}}{\partial y} =0 \qquad &y=0		
	\label{eq_neumann5}
	\\		 
	\frac{\partial \phi_{\pm,l}}{\partial y} =\frac{\partial \phi_{\pm,s}}{\partial y} 
   	=0 \qquad &y=H	
   	\label{eq_neumann6}		
	\\
		\frac{\partial c_2}{\partial y}	=  \frac{\partial c_4}{\partial y}=0
		\qquad &y=H
	\label{eq_neumann7}
\end{align}

\begin{table}[htbp]
\small
	\newcommand{\tabincell}[2]{\begin{tabular}{@{}#1@{}}#2\end{tabular}}
	\centering
	\caption{Default values for 2D mathematical model for vanadium redox flow battery.}
	\begin{tabular}{clcc}
		\toprule
		 Symbol &      Quantity & Value  & Unit   \\
		\midrule
		$H,L,W$  & Cell height, thickness, width & $50, 3.28, 20$  & $10^{-3}$m \\
		$a$ & Electrode specific area & 57622 &m$^2$/m$^3$ \\				
		$d_m$  & Membrane thickness & $5.08 \times 10^{-5} $ & m \\
		$I$ & Total current  & 2 &  A\\
		\makecell{$D_2, D_4, D_H^+$, \\    $D_{SO_4^{2-}},D_{HSO_4^{-}}$} &
		 Diffusion coefficient & \makecell{2.4, 3.9, 93.12, \\ 10.65, 13.3} & $10^{-10}$m$^2$/s\\
		$\vec{v}$ & Flow velocity & (0, $5.08 \times 10^{-3}$) &m/s \\
		$T$ & Cell temperature & 293.15 &K \\
		$\sigma_{s}, \sigma_m$ & Electrode, membrane  conductivity & 500, 30 &S/m \\
		$E_{+}^0, E_{-}^0$ & Standard potential & 1.004, -0.255 &V \\
		$k_{+}, k_{-}$ & Standard reaction rate constant & 1.1, 3.0  & $10^{-6}$m/s\\
		$\alpha_{\pm, c}, \alpha_{\pm, a}$ & Transfer coefficient & 0.5 & 1 \\
		$c_{+,H^+}^0$ & Initial $H^+$ concentration & $7000 + 3000 \times SOC$ & mol/m$^3$\\
		$c_{-,H^+}^0$ & Initial $H^+$ concentration & 5500  & mol/m$^3$\\
		$c_{\pm,HSO_4^-}^0$ & Initial $HSO_4^-$ concentration & 2500 & mol/m$^3$ \\
		$c_{+,H_2O}^0$ &Initial positive side $H_2O$ concentration & $30000-1500\times SOC$ & mol/m$^3$ \\		
		$c_{-,H_2O}^0$ &Initial negative side $H_2O$ concentration & 46100  & mol/m$^3$ \\
		$\epsilon$ & Electrode porosity & 0.92317 & 1\\
		$c^0$ & Initial vanadium concentration & 1500 & mol/m$^3$ \\
		\bottomrule
	\end{tabular}
	\label{table_params}
\end{table}

\subsubsection{Summary}
For the sake of clarity, the governing equations for the 2D mathematical model, comprising of Eqs. \eqref{eq_c2}, \eqref{eq_c4}, \eqref{eq_phil} and \eqref{eq_phis}, are summarized as
\begin{align}
	\vec{v} \cdot \nabla c_2 -D_2 \nabla^2 c_2 &= -j_{-, tot}/F  &
	\label{eq_gov_1}
	\\
	-\sigma_{-,l}^{\text{eff}}\nabla^2 \phi_{-,l} &=  j_{-, tot} 
	&\qquad (x,y, SOC) \in \Omega _- \times \mathcal{T}
	\label{eq_gov_2}
	\\
	-\sigma_{-,s}^{\text{eff}} \nabla^2 \phi_{-,s} &= -j_{-,tot} 	&
	\label{eq_gov_3}
	\\
	\vec{v} \cdot \nabla c_4 -D_4 \nabla^2 c_4 &= -j_{+, tot}/F  &
	\label{eq_gov_4}
	\\
	-\sigma_{+,l}^{\text{eff}}\nabla^2 \phi_{+,l} &=  j_{+, tot} 
	&\qquad (x,y, SOC) \in \Omega _+ \times \mathcal{T}
	\label{eq_gov_5}
	\\
	-\sigma_{+,s}^{\text{eff}} \nabla^2 \phi_{+,s} &= -j_{+,tot} 	&	
	\label{eq_gov_6}
\end{align}
where $\Omega _-=[-L, 0] \times [0,H]$, $\Omega _+=[0, L] \times [0,H]$ and $\mathcal{T} = [SOC_{\min}, SOC_{\max}]$. For the present problem, we set the range of SOC as $[SOC_{\min}, SOC_{\max}] = [0.1, 0.8] $. The variables to be solved are $\{c_2, \phi_{-,l}, \phi_{-,s}, c_4, \phi_{+,l}, \phi_{+,s}\}$. Eq. \eqref{eq_membrane} for the membrane can be deemed as a special kind of boundary condition. Therefore, the present model has 7 boundaries, containing 24 boundary conditions.

\subsection{Physics-informed machine learning}
\subsubsection{Network architecture}
\begin{figure}[H]
	\centering	
	\includegraphics[width=12cm]{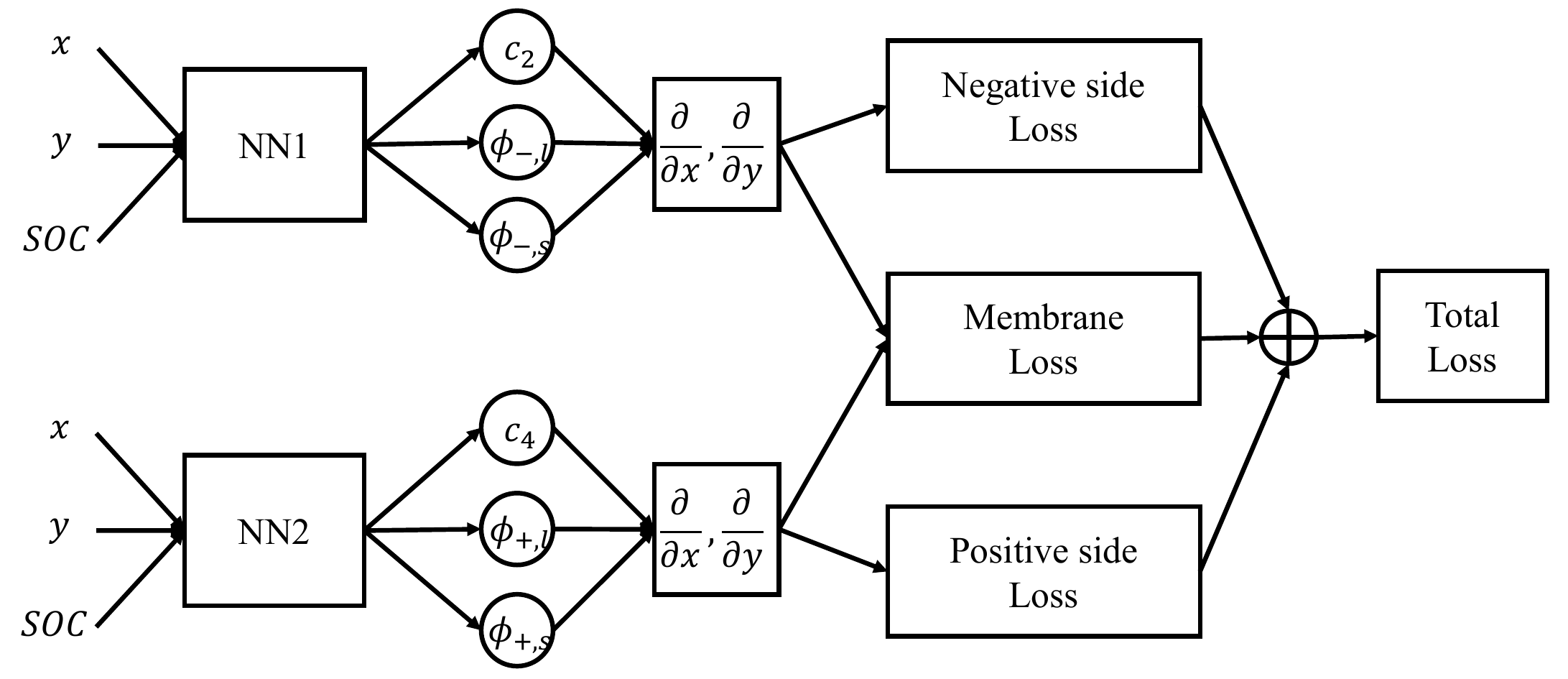}
	\caption{The neural network architecture and the forward propagation process for building the loss function.}
	\label{fig_net}	
\end{figure}

For the aforementioned 2D  mathematical model, the unknowns on the negative cell side ($c_2, \phi_{-,l}, \phi_{-,s}$) and  the unknowns on the positive cell side ($c_4, \phi_{+,l}, \phi_{+,s}$) are strongly coupled, so they need to be solved together. The discontinuity of the electrolyte/electrode potential across the membrane interface poses a challenge when trying to approximate it with a single neural network output across the entire domain. Therefore,  we build a composite network containing two sub-networks to approximate the unknowns on the negative and positive sides, respectively. As shown in Fig. \ref{fig_net}, the input of both sub-networks is $(x,y,SOC)$, the output of the first network is ($c_2, \phi_{-,l}, \phi_{-,s}$), and the output of the second sub-network is ($c_4, \phi_{+,l}, \phi_{+,s}$). As for the  architecture of each sub-network, we choose to use the modified feedforward neural network (FNN) \cite{wang2021understanding}, which outperforms the traditional FNN. The modified FNN consists of an input layer, $L$ hidden layers with each layer containing $n_l$ neurons and an output layer. The forward propagation of the modified FNN is defined as follows:
\begin{equation}
	\left\{
	\begin{aligned}
		\mathbf{U}\;&= f_{act}(\mathbf{W}^U\mathbf{x}+\mathbf{b}^U) &&\\
		\mathbf{V}\;&= f_{act}(\mathbf{W}^V\mathbf{x}+\mathbf{b}^V) &&\\
		\mathbf{y}^{1} &= f_{act}( \mathbf{W}^{1}\mathbf{x}+ \mathbf{b}^{1}) &&\\
		\mathbf{Z}^{l} &= f_{act}( \mathbf{W}^{l}\mathbf{y}^{l-1}+ \mathbf{b}^{l}), && 2\le l \le L \\
		\mathbf{y}^{l} &= (1-\mathbf{Z}^{l}) \odot \mathbf{U} + \mathbf{Z}^{l} \odot \mathbf{V}, && 2\le l \le L \\
		\mathbf{y} \;     &=\mathbf{y}^{L+1}=\mathbf{W}^{L+1}\mathbf{y}^{L}+ \mathbf{b}^{L+1} &&\\
	\end{aligned}
	\right .
	,
\end{equation}
where $\mathbf{x} \in \mathbb{R}^{n_0}$ is the input, $\odot$ denotes a point-wise multiplication, $f_{act}$ is a point-wise activation function. $\mathbf{y}^{l}  \in \mathbb{R}^{n_{l}} $ is the output of the $l_\text{th}$ layer. $\mathbf{W}^{l} \in \mathbb{R}^{n_{l}} \times \mathbb{R}^{n_{l-1}}$  and $\mathbf{W}^U \in \mathbb{R}^{n_1} \times \mathbb{R}^{n_0} $, and $ \mathbf{W}^V \in \mathbb{R}^{n_1} \times \mathbb{R}^{n_0}$ are the weights.
$\mathbf{b}^{l} \in \mathbb{R}^{n_{l}} $, $\mathbf{b}^{U} \in \mathbb{R}^{n_1}$ and  $\mathbf{b}^{V} \in \mathbb{R}^{n_1}$ are the biases. 

% \vspace{-0.5cm}

\subsubsection{Normalization}
\label{sec_norm}
The training performance of neural networks is usually affected by the magnitude scale of the input/output \cite{loffe2014accelerating}, thus normalization before  the training process can help. 
The minimum and maximum bound of the input $\mathbf{x}=(x,y,SOC)$ is predefined, and thus can be used to scale the input $\mathbf{x}$ to $[-1, 1]^3$ with a simple linear transformation.
For the output of the first sub-network, we have prior knowledge that the output component $c_2$ should lie in the range $[c_2^{in}, c^0]$ for the charging stage and  $[0, c_2^{in}]$ for the discharging stage. For this purpose, we build a smooth transformation as follows:
\begin{equation}
	c_2 = c^0 \left[ \frac{SOC+charge}{2} + \frac{SOC-charge}{2}\sin \left(\frac{\widehat{c_2} \pi}{2} \right) \right]
\end{equation}
where $charge=1/0$ denotes the charging/discharging stage. Note that $\widehat{c_2}$ is the raw value of $c_2$ output by the first sub-network.
 Analogously, the output component $\widehat{c_4}$ of the second sub-network is transformed as:
\begin{equation}
	c_4 = c^0 \left[ \frac{2-SOC-charge}{2} + \frac{SOC-charge}{2}\sin \left(\frac{\widehat{c_4} \pi}{2} \right) \right]
\end{equation}
The output components ${\phi_{\pm,l}}$ or ${\phi_{\pm,s}}$ are normalized with the self-defined ranges (given in Table \ref{table_range_out}) as follows: 
\begin{align}
	\phi_{\pm,l} &= \frac{\phi_{\pm,l}^{\min}+\phi_{\pm,l}^{\max}}{2}+\frac{\phi_{\pm,l}^{\min}-\phi_{\pm,l}^{\max}}{2}\widehat{{\phi_{\pm,l}}} \\
	\phi_{\pm,s} &= \frac{\phi_{\pm,s}^{\min}+\phi_{\pm,s}^{\max}}{2}+\frac{\phi_{\pm,s}^{\min}-\phi_{\pm,s}^{\max}}{2}\widehat{{\phi_{\pm,s}}} 
\end{align}
Note that these ranges should be selected carefully according to battery properties.
\begin{table}[tbp]
	\newcommand{\tabincell}[2]{\begin{tabular}{@{}#1@{}}#2\end{tabular}}
	\centering
	\caption{The self-defined ranges for the normalization of potentials.}
	\begin{tabular}{ccccc}
		\toprule
		Stage &Symbol &    (min, max)  & Symbol & (min, max)   \\
		\midrule
		\multirow{2}{*}{ charging} &$\phi_{-,l}$ & $(0.25,0.60)$   & $\phi_{-,s}$ & $(0, 0.30)$  \\              
		& $\phi_{+,l}$ & $(0.30,0.60)$   & $\phi_{+,s}$ & $(1.40,2.20)$  \\
		\multirow{2}{*}{ discharging} &$\phi_{-,l}$ & $(-0.20,0.25)$   & $\phi_{-,s}$ & $(-0.1, 0)$  \\              
		& $\phi_{+,l}$ & $(-0.25,0.20)$   & $\phi_{+,s}$ & $(0.30,1.50)$  \\		
		\bottomrule
	\end{tabular}
	\label{table_range_out}
\end{table}

In addition, the governing equations   \eqref{eq_gov_1}-\eqref{eq_gov_6} and boundary conditions  \eqref{eq_membrane}-\eqref{eq_neumann7} are at different orders of magnitude. This can significantly hinder  the solving capability of PINNs. Here, we  nondimensionalize each of the governing equations or boundary conditions by dividing it with a constant coefficient, which is listed in Table \ref{table_normal_coeff}.

During the training process, the calculation of the total current density through Eqs. \eqref{eq_jtot1} and \eqref{eq_jtot2} involves the exponential of the overpotential $\eta_{\pm}$. This may induce an extremely large value of total current density, thus leading to a gradient exploding problem. To overcome this issue, we explicitly restrict the value of $\eta_{\pm}$ within the range of [-0.1, 0.1] by the formulation:
\begin{equation}
	\eta_{\pm} = \max\left( \min\left(\widehat{\eta_{\pm}}, 0.1 \right), -0.1 \right)
\end{equation}
where $\widehat{\eta_{\pm}}$ is the raw value of $\eta_{\pm}$ estimated by the sub-network outputs.

\begin{table}[tbp]
	\newcommand{\tabincell}[2]{\begin{tabular}{@{}#1@{}}#2\end{tabular}}
	\centering
	\caption{The normalization coefficients for the governing equations and boundary conditions. Here, $\beta=1-SOC$ for the charging stage and $\beta=SOC$ for the discharging stage.}
	\begin{tabular}{ccccc}
		\toprule
		Equations &coefficient &  Equations &coefficient   \\
		\midrule
		\eqref{eq_gov_1}, \eqref{eq_gov_4}  & $\beta c^0\|\vec{v}\|_2/H$ &
		\eqref{eq_gov_2}, \eqref{eq_gov_3},
		\eqref{eq_gov_5}, \eqref{eq_gov_6} & $\beta Fc^0\|\vec{v}\|_2/H$
		\\
		\eqref{eq_membrane}, \eqref{eq_i_avg}, \eqref{eq_bc_membrane} & $i_{avg}$ & \eqref{eq_inlet_c2}, \eqref{eq_inlet_c4} & $c^0$ \\
		\eqref{eq_0_phi_s} & 1 & \eqref{eq_neumann1}, \eqref{eq_neumann3} & $c^0/L$ \\
		\eqref{eq_neumann2}, \eqref{eq_neumann4} & $i_{avg}$ &
		\eqref{eq_neumann5}, \eqref{eq_neumann6} & $1/H$ \\
		\eqref{eq_neumann7} & $c^0/H$ \\
		\bottomrule
	\end{tabular}
	\label{table_normal_coeff}
\end{table}

\subsubsection{Loss function}
For the present problem, the loss function is defined as 
\begin{equation}
	\label{eq_loss_forward}
	\begin{aligned}
		\mathcal{L}(\pmb{\theta})&= 
			\sum_{j=1}^{N_{op}}{\left(\frac{1}{N_{R,j}}\sum_{i=1}^{N_{R,j}}{\left\|\mathcal{N}_j(\mathbf{y}(\mathbf{x}_{R,j}^i))\right\|^2}\right)}
			+
		\frac{1}{N_{D}}\sum_{i=1}^{N_{D}}{\left\|\mathbf{y}\left(\mathbf{x}_{D}^i\right)-\mathbf{y}^{i,*}\right\|^2} 
	\end{aligned}
	,
\end{equation} 
where $N_{op}=6+24=30$, $\mathcal{N}_j$ denotes the differential/boundary operators defined by the governing equations \eqref{eq_gov_1}-\eqref{eq_gov_6} or boundary conditions \eqref{eq_membrane}-\eqref{eq_neumann7}, respectively. $N_{R,j}$ denotes the size of  the residual data set $\mathcal{D}_{R,j}=\{\mathbf{x}_{R,j}^i\}_{i=1}^{N_{R,j}}$ for the j$th$ differential/boundary operator. $N_{D}$ denotes the size of  the labeled data set $\mathcal{D}_{ D}=\{(\mathbf{x}_{D}^i, \mathbf{y}_{ D}^{i,*})\}_{i=1}^{N_{D}}$ (if any are available).

\subsubsection{Training with self-adaptive weights}
\label{subsec_training}
The 2D mathematical model is comprised of 6 differential equations and 24 boundary conditions. Although we have normalized these equations roughly to the order of 1 by applying a coefficient to each equation, it is still necessary to further balance these terms with adaptive weights. We adopt a recently proposed self-adaptive weighting method  \cite{mcclenny2020self} during the training process.

By assigning a weight to each individual training point, the loss function in Eq. \eqref{eq_loss_forward} is recast as 
\begin{equation}
	\label{eq_lossMF_SA}
	\begin{aligned}
		\mathcal{L}(\pmb{\theta}, \mathbf{w}) 
		&= \sum_{j=1}^{N_{op}}{\left(
			\frac{1}{N_{R,j}}\sum_{i=1}^{N_{R,j}}{M( w_{R,j}^i)\left\|\mathcal{N}_j(\mathbf{y}(\mathbf{x}_{R,j}^i))\right\|^2}\right)}
		\\
		&+
		{
			\frac{1}{N_{D}}\sum_{i=1}^{N_{D}}{M(w_{D}^i)\left\|\mathbf{y}(\mathbf{x}_{D}^i)-\mathbf{y}_{D}^{i,*}\right\|^2}
		}  \\
	\end{aligned}	
\end{equation}
where $\mathbf{w}$ are the collection of the trainable weights and $M$ is a mask function. In this paper, we choose the mask function $M(x)=x^2$, which is also used in \cite{mcclenny2020self}. The loss function in Eq. \eqref{eq_lossMF_SA} is minimized with respect to the parameters $\pmb{\theta}$, but maximized with respect to the weights $\mathbf{w}$, namely
\begin{equation}
	\min_{\pmb{\theta}} \max_{\mathbf{w}}  \mathcal{L}(\pmb{\theta}, \mathbf{w}).
\end{equation}
Following a gradient ascent/descent approach, the parameters and weights are updated concurrently, namely
\begin{equation}
	\begin{aligned}
		\pmb{\theta}^{k+1}  &= \pmb{\theta}^{k} - \eta^k \nabla_{\pmb{\theta}}\mathcal{L}(\pmb{\theta}^k, \mathbf{w}^k) \\
		\mathbf{w}^{k+1} &= \mathbf{w}^{k} + \rho^k \nabla_{\mathbf{w}}\mathcal{L}(\pmb{\theta}^k, \mathbf{w}^k) \\
	\end{aligned} 
\end{equation}
where $k$ is the iteration number, while $\eta^k$ and $\rho^k$ are the learning rates for the parameters and self-adaptive weights, respectively. The gradient with respect to the self-adaptive weights is calculated as follows
\begin{equation}
	\begin{aligned}
		\nabla_{\mathbf{w}}\mathcal{L}(\pmb{\theta}, \mathbf{w})
		&=
		\left\{
		\left\{
		{M^{\prime}(w_{R,j}^i)\left\|\mathcal{N}_j(\mathbf{y}(\mathbf{x}_{R,j}^i))\right\|^2} 
		\right\}_{i=1}^{N_{R,j}}
		\right\}_{j=1}^{N_{op}}             \\
		&\cup
		\left\{
		{M^{\prime}(w_{D}^i)\left\|\mathbf{y}(\mathbf{x}_{D}^i)-\mathbf{y}_{D}^{i,*}\right\|^2} 	\right\}_{i=1}^{N_{D}},      
	\end{aligned} 	\label{eq_gradien_SA2}
\end{equation}
where $M'$ stands for the derivative of the mask function $M$ with respect to its argument. Note that the above gradients can be calculated directly without  automatic differentiation.

\subsubsection{Training details }
Each sub-network is constructed with 6 hidden layers and each hidden layer contains 50 neurons. The activation function used in these networks is the Swish function given by $f_{act}(x)=x\cdot \text{sigmoid}(x)$ \cite{ramachandran2017searching}. To initialize the network, we use the Xavier initialization method \cite{glorot2010understanding} for the weights and set the biases to 0. The self-adaptive weights are initially set to 1. The training of the network takes place in two stages. First, the Adam optimizer \cite{kingma2014adam} is used for 36,000 iterations. This is followed by 4,000 iterations using the L-BFGS optimizer \cite{liu1989limited}. During the Adam optimization stage, the initial learning rate $\eta$ for the parameters $\pmb{\theta}$ is set at 0.001 and the learning rate decreases by 1\% every 200 iterations. At this stage, the self-adaptive weights are also updated with a fixed learning rate $\rho=0.1$. During the second stage of training with the L-BFGS optimizer, the self-adaptive weights remain constant and are not updated. 
The training process for the corresponding neural networks is carried out using PyTorch \cite{paszke2019pytorch}, and runs on a server-grade supercomputer, utilizing 32-bit single-precision data type and operatering with a single GPU (NVIDIA$^\circledR$ Tesla P100).

The charging and discharging stages are trained independently with the same network setup.  For training of each stage, 10000 residual points are randomly chosen for each half-cell side, 1800 boundary points are randomly chosen for each vertical boundary, 200 boundary points are randomly chosen for each horizontal boundary.

\section{Results and discussion}
\begin{figure}[ht]
	\centering	
	\includegraphics[width=15.5cm]{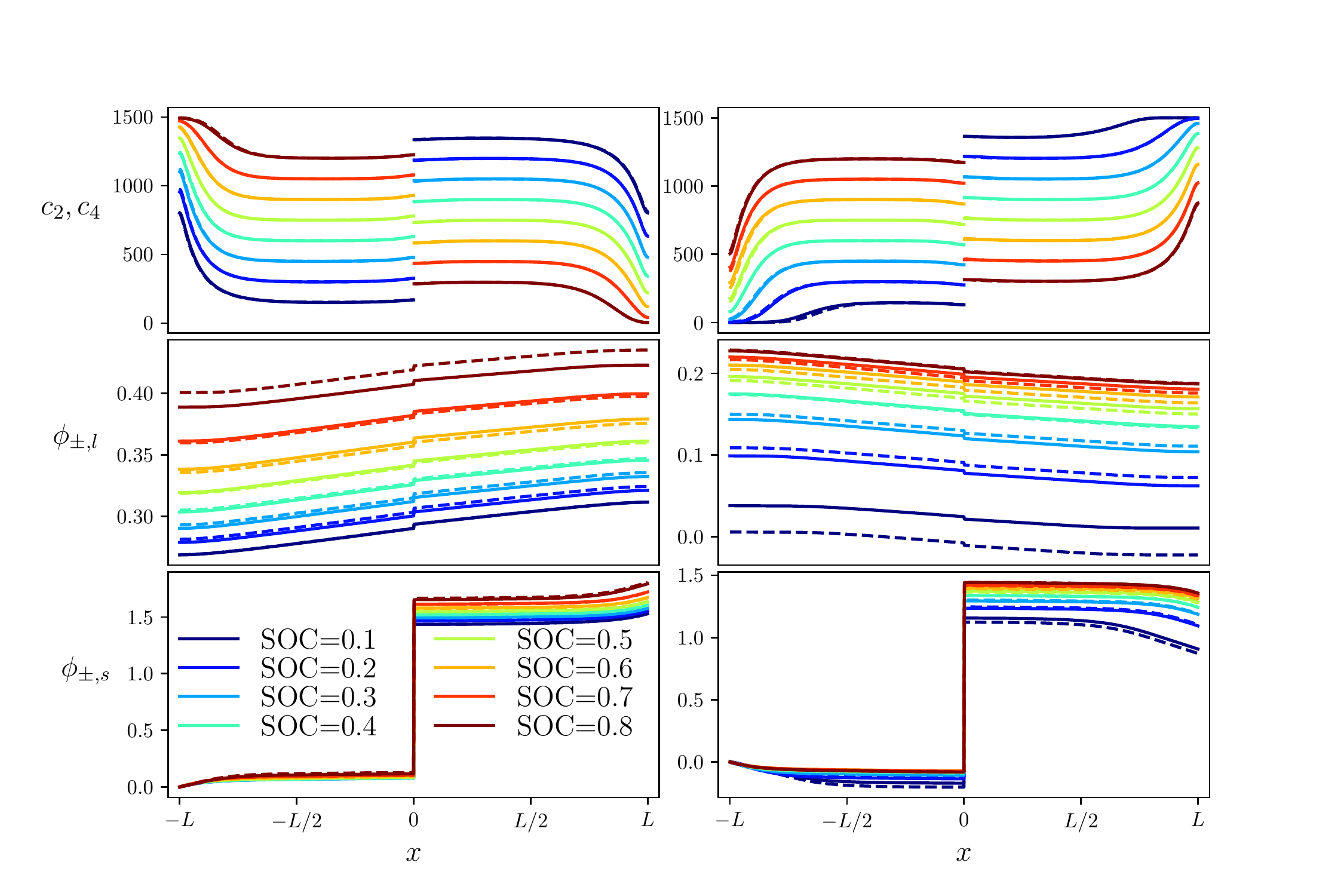}
	\caption{Predictions at the outlet $y=H$ for the charging stage (left) and discharging stage (right) of PINN (soild lines)  and COMSOL (dashed lines) for the total current $I=2$A.}
	\label{fig_PINN_2A}	
\end{figure}

To validate the prediction accuracy of the PINN approach, we also solved the 2D mathematical model with COMSOL$^\circledR$ based on finite element method \cite{dickinson2014comsol}. The solutions at the outlet $y=H$ predicted from COMSOL and PINN  are shown in Fig. \ref{fig_PINN_2A}. Comparison of predictions from COMSOL and PINN shows a constant-like potential shift for each $SOC$ value. The shift magnitude of the electrolyte potential is very similar to that of the electrode potential. This arises due to the nature of the 2D mathematical model, where the electrolyte potential only appears in derivative form or as a difference with the electrode potential.
The electrode potential shift primarily originates from the underestimated derivative of the electrode potential at the negative current collector, implying the underestimation of current density at the negative current collector. 

\begin{figure}[ht]
	\centering	
	\includegraphics[width=15.5cm]{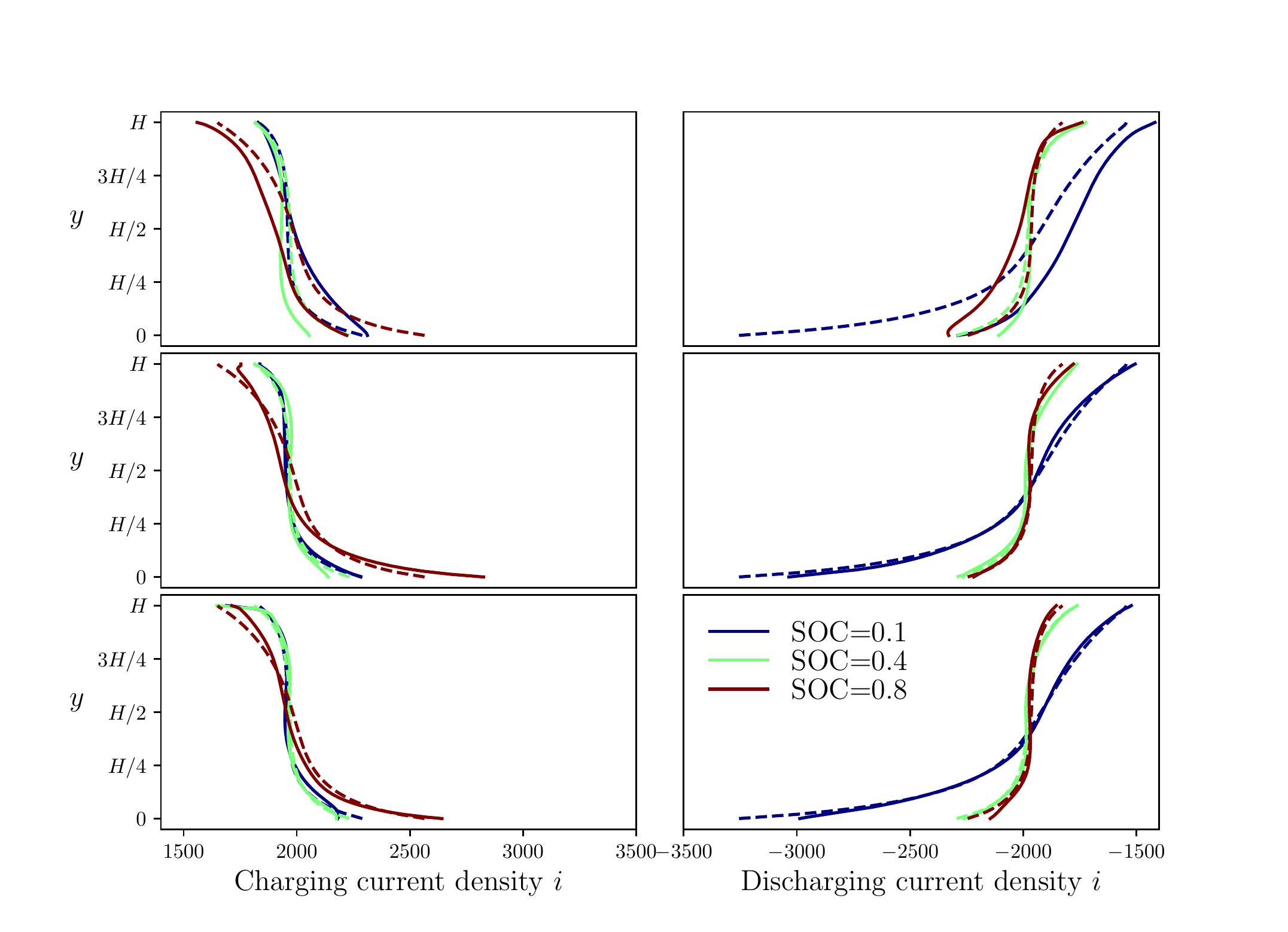}
	\caption{Predictions of current density from the PINN (top), EPINN (middle) and EPINN with data (bottom) for the charging (left) and discharging (right) stages. Solid lines represent the output of the PINNs, while dashed lines indicate COMSOL's results. These predictions are carried out for the total current $I=2$A." }
	\label{fig_current}	
\end{figure}

Figure \ref{fig_current} illustrates the predicted current density at the negative current collector from both the PINN and COMSOL. It's evident that the PINN underestimates the current density at the end of both the charging stage ($SOC=0.8$) and the discharging stage ($SOC=0.1$).
To further understand this phenomenon, we will examine the discharging stage. At the end of the discharging stage, the species V$^{2+}$ at the outlet is nearly depleted, which substantially weakens the negative reaction in Eq. \eqref{eq_reaction_neg}. As a result, the  current density at the outlet falls far below the average value. To maintain a consistent average value over the negative collector, the current density at the inlet is correspondingly higher. This sharp change in current density from the inlet to the outlet presents a considerable challenge for  the PINN training process.

% \vspace{-0.5cm}

\subsection{Predictions with additional prior knowledge}
The PINN's underestimation of current density violates the conservation of current,  that is,  the total current over negative collector and membrane  is equal to that at positive collector. Since the governing equations defined in  Section \ref{sec_2Dmodel} are well defined, the conservation of current is implicitly constrained by the governing equations. However, PINN is unable to resolve the sharp changes of current density. To remedy this situation, we explicitly constrain the PINN  to conserve the total current:
\begin{equation}
		\int_{0}^{H} \sigma_{-,s}^{\text{eff}} \frac{\partial 	\phi_{-,s}(-L,y,SOC)}{\partial x} dy 
	= 	\int_{0}^{H} \sigma_{-,l}^{\text{eff}} \frac{\partial 	\phi_{-,l}(0,y,SOC)}{\partial x} dy =\pm i_{avg}H
	\label{eq_add_physics}
\end{equation}

\begin{figure}[ht]
	\centering	
	\includegraphics[width=15.5cm]{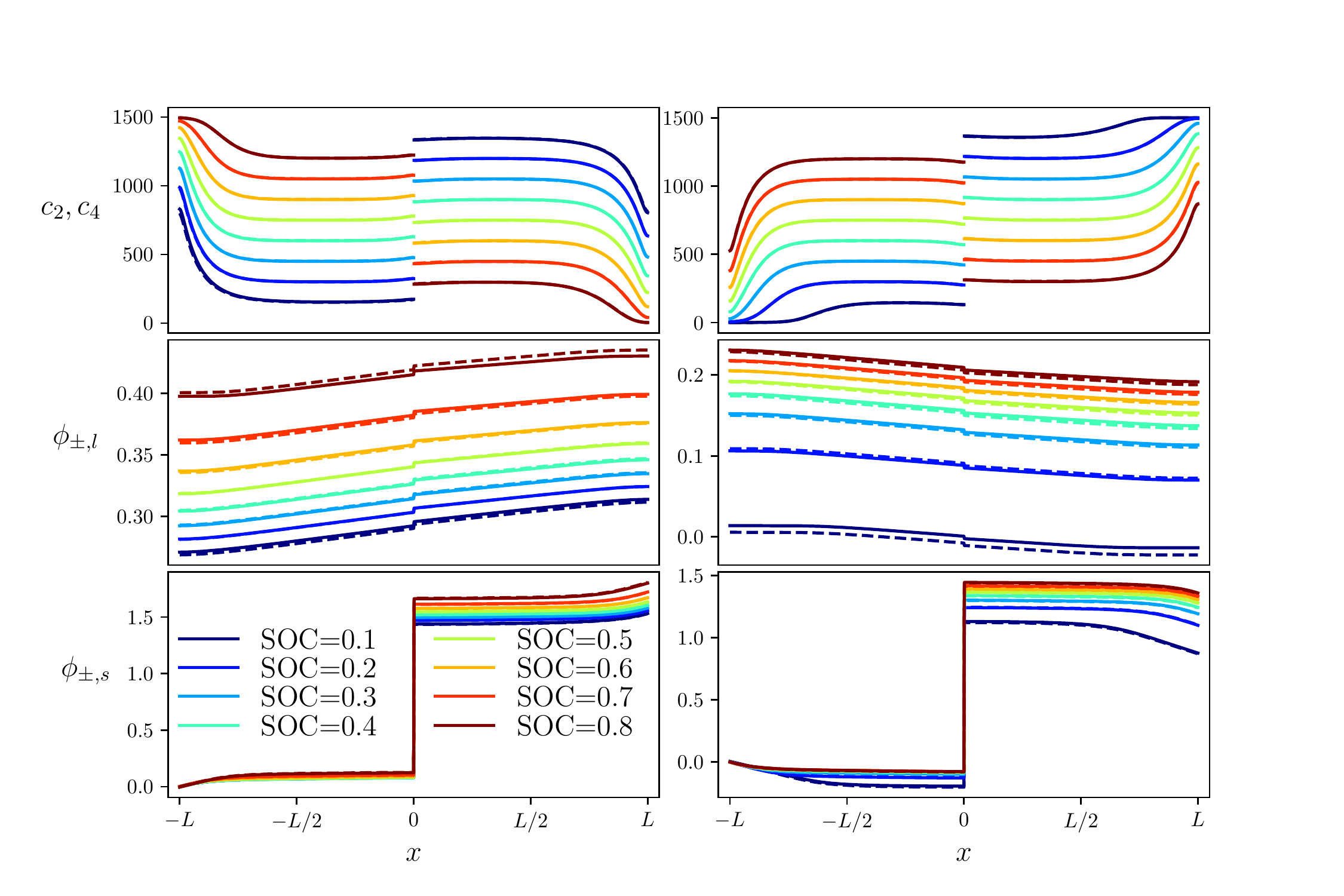}
	\caption{Predictions at the outlet $y=H$ for the charging stage (left) and discharging stage (right) of the enhanced PINN (soild lines)  and COMSOL (dashed lines) for the total current $I=2$A.}
	\label{fig_EPINN_2A}	
\end{figure}

We enforce  Eq. \eqref{eq_add_physics} by introducing another term in the loss function. To evaluate, we generate 101 uniformly-distributed $SOC$ values, and the corresponding integral over $y$ is approximated by the quadrature integral over 101 uniformly-distributed points for each $SOC$ value. We refer to the  PINN trained with the additional constraint as ``enhanced PINN" or ``EPINN".
The current density at the negative collector is shown in Fig. \ref{fig_current}. It is shown that
the discrepancy between current density predicted by EPINN and COMSOL  is significantly reduced. Meanwhile, as shown in Fig. \ref{fig_EPINN_2A}, the constant-like shift of potentials is also reduced substantially, and the small deviation of species concentration shown in Fig. \ref{fig_PINN_2A} is also fixed.

% \vspace{-0.5cm}

\begin{figure}[ht]
	\centering	
	\includegraphics[width=15.5cm]{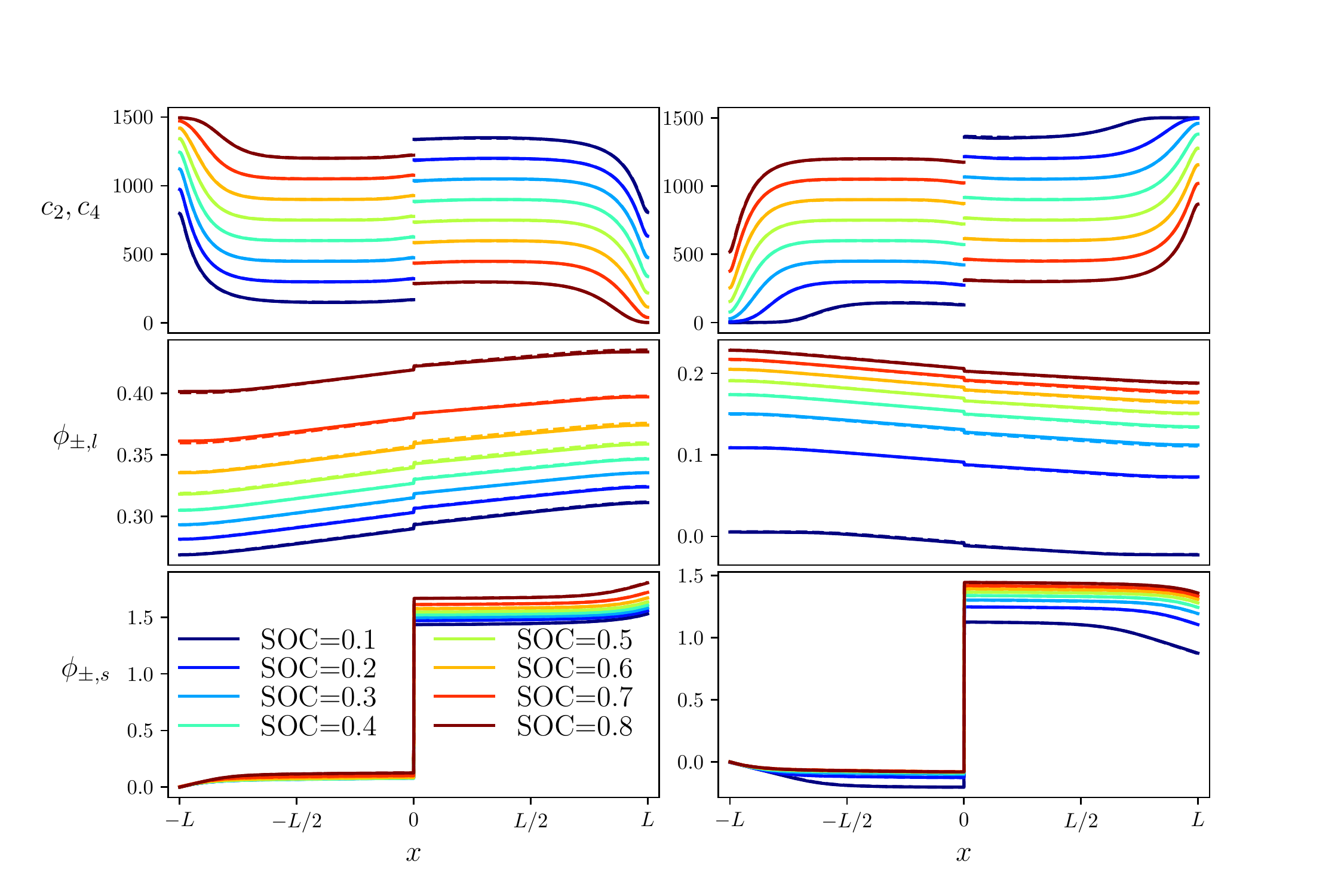}
	\caption{Predictions at the outlet $y=H$ for the charging stage (left) and discharging stage (right) of the enhanced PINN with data (soild lines)  and COMSOL (dashed lines) for the total current $I=2$A.}
	\label{fig_EPINND_2A}	
\end{figure}% \vspace{-0.5cm}

\begin{figure}[ht!]
	\centering	
	\subfigure[]{
	\includegraphics[width=7cm]{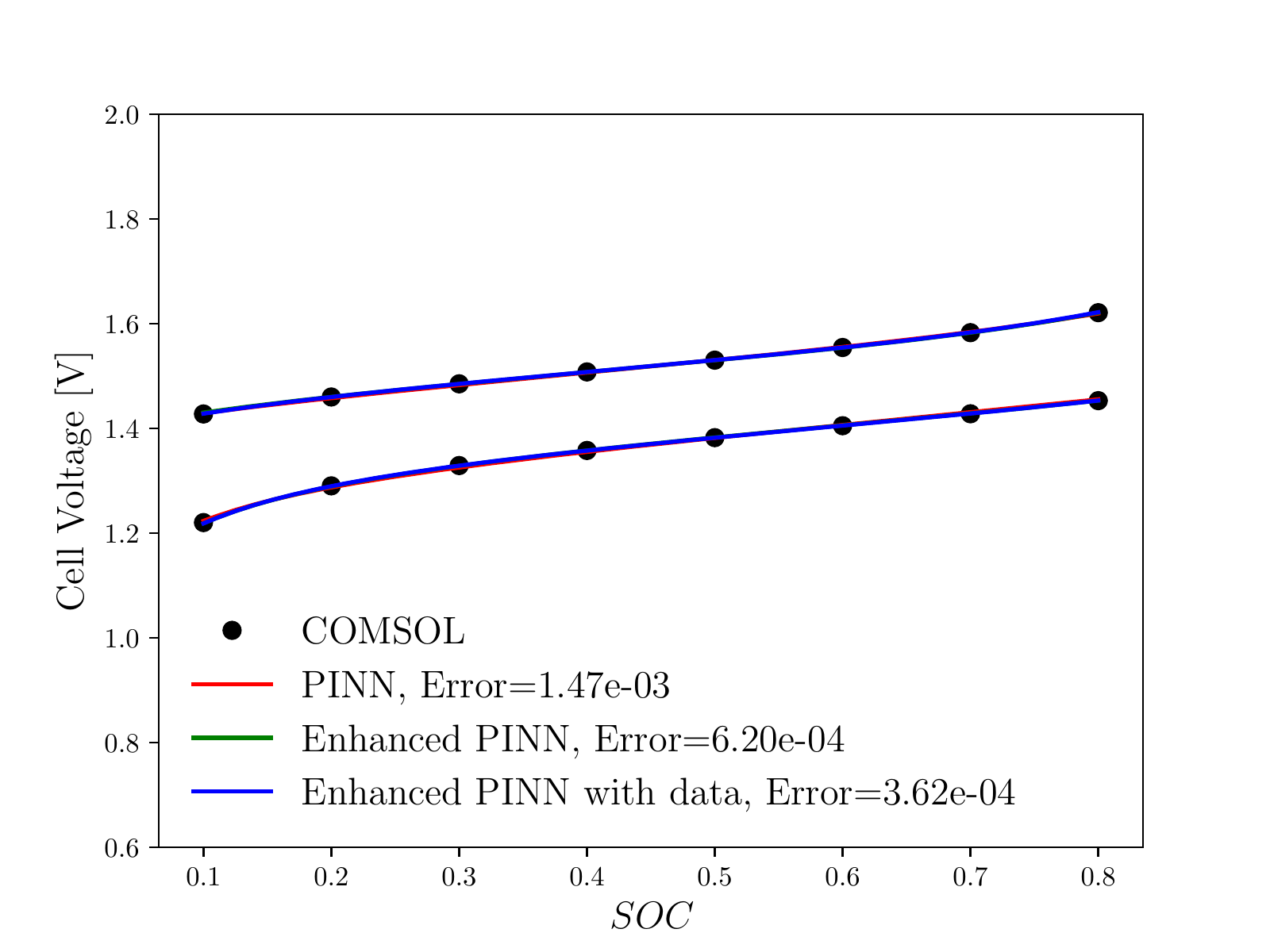}
	}	
	\subfigure[]{
	\includegraphics[width=7cm]{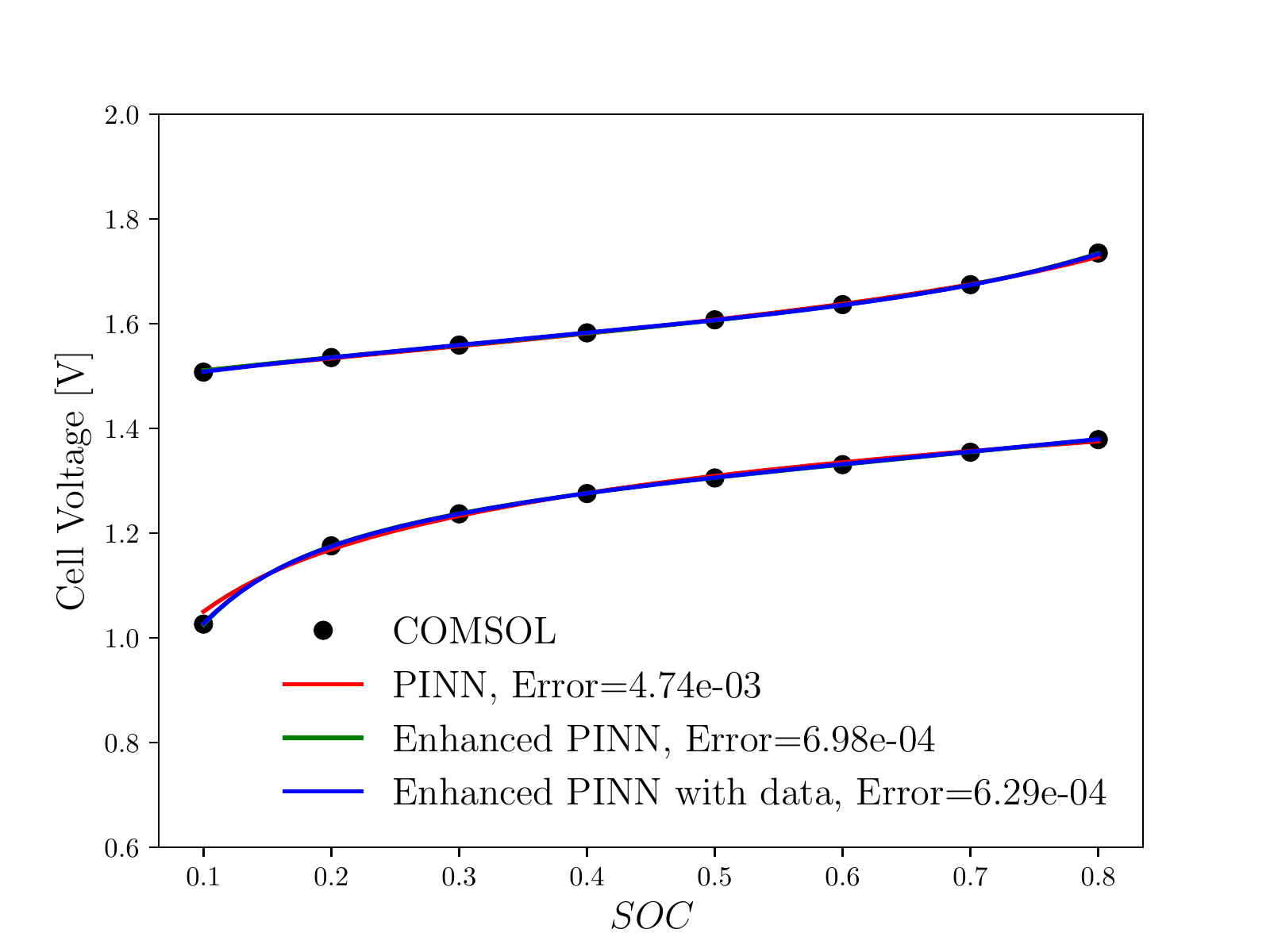}
}	
	\subfigure[]{
	\includegraphics[width=7cm]{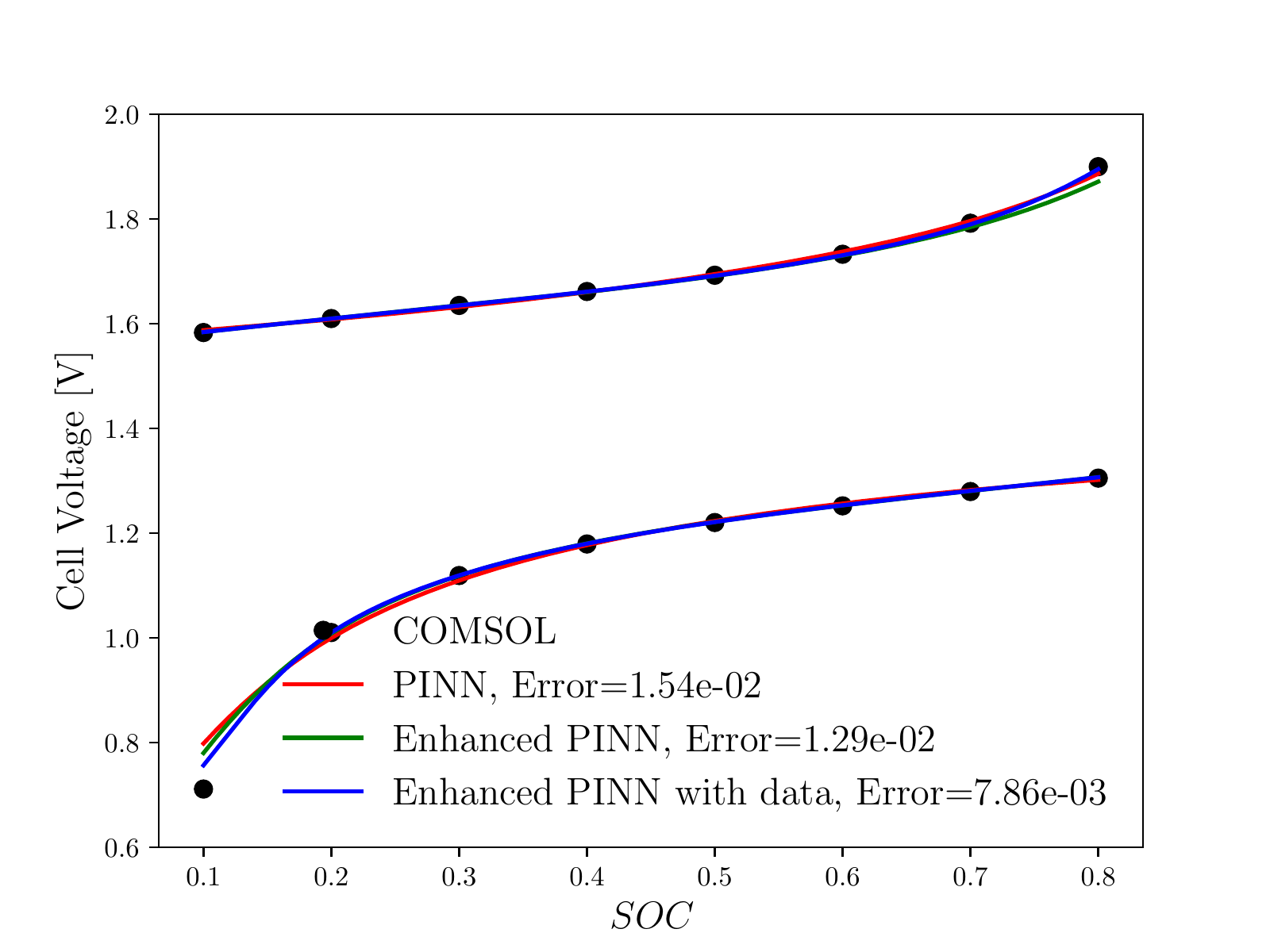}
}	
	\caption{Predictions of cell voltage for the total current $I=1$A (a), 2A (b) and 3A (c). The error in the legend denotes the relative $L_2$ error.}
	\label{fig_CV}	
\end{figure}

To make the predictions even more precise, we assume that a small amount of data is available, and we feed these data into the training of EPINN to investigate the influence on the prediction accuracy. To generate the data, we extract electrolyte potentials at 40 randomly sampled points from the membrane and  negative collector, respectively.
The predictions of the EPINN with data are shown in Fig. \ref{fig_EPINND_2A} and Fig. \ref{fig_CV}. It is shown that the constant-like shift is eliminated and prediction of solutions are in excellent agreement with those form COMSOL. This indicates that the performance of the EPINN improves in the presence of a small amount of data.

% \vspace{-0.5cm}

\subsection{Overall accuracy}
In Fig. \ref{fig_CV}, the cell voltage curves are  plotted for the total current $I=1, 2, 3$A. Despite the presence of a potential shift for the total current $I=2$A, the cell voltage predicted by the PINN aligns well with that of COMSOL, showcasing a tiny relative error. 
This can be attributed to the fact that the cell voltage is calculated as the difference between the electrode potentials at the positive and negative collectors. As a result, the constant-like shift does not significantly affect the overall accuracy. Even so, we can still observe an obvious accuracy improvement of the cell voltage prediction through  EPINN and EPINN with data by reducing the constant-like shift. In addition, it is shown that 
the relative error of the PINN, EPINN and EPINN with data increases with a larger magnitude of the total current. For $I=3$A,  significant deviations are also observed at the end of charging and discharging stages. This suggests that an increase in the total current will intensifie the sharpness of the change in current density at the negative current collector. 

The  computational cost of COMSOL simulations and the training of PINNs are given in Table \ref{table_cost}.
Relative to the PINN, the inclusion of additional terms in the loss function of the EPINN and EPINN with data incur only small amounts of extra training time. 
Despite the fact that these computations are executed on distinct devices, it is still reasonably valid to infer that the computational time associated with the training of PINNs markedly exceeds that of COMSOL simulations. Yet, an inherent advantage of PINNs resides in the decoupling of their training and evaluation procedures. After the termination of the training process, PINNs can instantaneously predict solutions at arbitrary points within the computation domain. In contrast, COMSOL's capacity for prediction is confined to a finite set of points, necessitating independent implementation for each $SOC$. To derive solutions beyond this fixed set, the deployment of a supplementary interpolation procedure is required. In addition, it is noted that the velocity of training for PINNs can be substantially augmented via established methodologies, such as the deployment of parallel computation employing multiple GPUs. {\color{black} In subsequent versions of the PINN approach, specific parameters from the 2D mathematical model in Section \ref{sec_2Dmodel}, such as the flow velocity and the total current, can also be integrated as additional inputs to the neural network. Once the network is trained, it will be capable of evaluating solutions at any spatial points and for various $SOC$ values, under a wide array of working conditions defined by these parameters.
}

\begin{table}[tbp]
	\newcommand{\tabincell}[2]{\begin{tabular}{@{}#1@{}}#2\end{tabular}}
	\centering
	\caption{Computational time cost and device setup for PINNs and COMSOL simulation. The cost of prediction is evaluated on $151\times201$ spatial points at $SOC=0.1, 0.2,..., 0.8$.}
	\begin{tabular}{ccc}
		\toprule
	 &Time (sec.) &  Device   \\
		\midrule
        Training of PINN & 43503 & \multirow{4}{*}{\makecell{NVIDIA Tesla P100 GPU (single)}} \\
        Training of EPINN & 45755  \\
		Training of EPINN with data & 48815  \\
        Prediction with trained PINNs & 0.48 &   \\ 
		Prediction with COMSOL & 216 & \makecell{Xeon Silver 4208 CPU, 16 cores}		\\
		\bottomrule
	\end{tabular}
	\label{table_cost}
\end{table}

\FloatBarrier
\section{Conclusions}
To investigate the application of physics-informed machine learning in the redox flow battery, we develop a physics-informed neural network approach for the predictions of unit cell vanadium redox flow battery based on a two-dimensional mathematical model. First, we simplify the three-dimensional physical model to a two-dimensional model based on several assumptions. Then we design a neural network consisting of two sub-networks to approximate the input-output mapping. To improve PINN's solving capability, we apply normalization to the input and output, nondimensionlise the governing equations and boundary conditions, and constrain the intermediate solutions (overpotentials). To further balance the different terms in the loss function, a self-adaptive weighing method is applied to the training process. The developed PINN is tested for a vanadium cell. First, the PINN with well-defined governing laws is able to predict the cell voltage correctly, but has difficulty in predicting the sharp change of current density at the end of charging and discharging stages. Second, an enhanced PINN (EPINN), further constraining the PINN with the conservation of current, outperforms the simple PINN in predicting both solution details and cell voltage. Third, if a small amount of experiment/simulation data is available, the prediction accuracy of the enhanced PINN can be further improved. Fourth, 
although the training time cost of PINNs far exceeds COMSOL simulation time cost, the prediction with trained PINNs is much more efficient than COMSOL prediction.

The present work provides a feasible PINN approach for the prediction of performance of a redox flow battery based on a complex mathematical model. The PINN approach has the potential for more complex scenarios, such as multi-query predictions.
Moreover, the PINN approach is expected to be further improved by integrating adaptive refinement approaches, or multi-fidelity approaches. 

\section*{Acknowledgments}
The authors would like to thank Dr. Jie Bao and Dr. Amanda Howard for very helpful discussions. 
The work is supported by the Energy Storage Materials
Initiative, a Laboratory Directed Research and Development Program at
Pacific Northwest National Laboratory (PNNL).
PNNL is a multi-program national laboratory operated for the U.S. Department of Energy (DOE) by Battelle Memorial Institute under Contract No. DE-AC05-76RL01830.

%\section*{References}
\bibliographystyle{elsarticle-num}
\bibliography{bibliography}

\begin{thebibliography}{10}
\expandafter\ifx\csname url\endcsname\relax
  \def\url#1{\texttt{#1}}\fi
\expandafter\ifx\csname urlprefix\endcsname\relax\def\urlprefix{URL }\fi
\expandafter\ifx\csname href\endcsname\relax
  \def\href#1#2{#2} \def\path#1{#1}\fi

\bibitem{soloveichik2015flow}
G.~L. Soloveichik, Flow batteries: current status and trends, Chemical reviews
  115~(20) (2015) 11533--11558.

\bibitem{noack2015chemistry}
J.~Noack, N.~Roznyatovskaya, T.~Herr, P.~Fischer, The chemistry of redox-flow
  batteries, Angewandte Chemie International Edition 54~(34) (2015) 9776--9809.

\bibitem{weber2011redox}
A.~Z. Weber, M.~M. Mench, J.~P. Meyers, P.~N. Ross, J.~T. Gostick, Q.~Liu,
  Redox flow batteries: a review, Journal of applied electrochemistry 41 (2011)
  1137--1164.

\bibitem{shah2008dynamic}
A.~Shah, M.~Watt-Smith, F.~Walsh, A dynamic performance model for redox-flow
  batteries involving soluble species, Electrochimica Acta 53~(27) (2008)
  8087--8100.

\bibitem{leung2017recent}
P.~Leung, A.~Shah, L.~Sanz, C.~Flox, J.~Morante, Q.~Xu, M.~Mohamed, C.~P.
  De~Le{\'o}n, F.~Walsh, Recent developments in organic redox flow batteries: A
  critical review, Journal of Power Sources 360 (2017) 243--283.

\bibitem{kim20131}
S.~Kim, E.~Thomsen, G.~Xia, Z.~Nie, J.~Bao, K.~Recknagle, W.~Wang,
  V.~Viswanathan, Q.~Luo, X.~Wei, et~al., 1 kw/1 kwh advanced vanadium redox
  flow battery utilizing mixed acid electrolytes, Journal of Power Sources 237
  (2013) 300--309.

\bibitem{skyllas2019performance}
M.~Skyllas-Kazacos, Performance improvements and cost considerations of the
  vanadium redox flow battery, ECS Transactions 89~(1) (2019) 29.

\bibitem{sum1985study}
E.~Sum, M.~Skyllas-Kazacos, A study of the v (ii)/v (iii) redox couple for
  redox flow cell applications, Journal of Power sources 15~(2-3) (1985)
  179--190.

\bibitem{skyllas1987efficient}
M.~Skyllas-Kazacos, F.~Grossmith, Efficient vanadium redox flow cell, Journal
  of the Electrochemical Society 134~(12) (1987) 2950.

\bibitem{kear2012development}
G.~Kear, A.~A. Shah, F.~C. Walsh, Development of the all-vanadium redox flow
  battery for energy storage: a review of technological, financial and policy
  aspects, International journal of energy research 36~(11) (2012) 1105--1120.

\bibitem{shah2011dynamic}
A.~Shah, R.~Tangirala, R.~Singh, R.~Wills, F.~Walsh, A dynamic unit cell model
  for the all-vanadium flow battery, Journal of the Electrochemical society
  158~(6) (2011) A671.

\bibitem{sharma2015verified}
A.~K. Sharma, C.~Ling, E.~Birgersson, M.~Vynnycky, M.~Han, Verified reduction
  of dimensionality for an all-vanadium redox flow battery model, Journal of
  Power Sources 279 (2015) 345--350.

\bibitem{eapen2019low}
D.~E. Eapen, S.~R. Choudhury, R.~Rengaswamy, Low grade heat recovery for power
  generation through electrochemical route: Vanadium redox flow battery, a case
  study, Applied Surface Science 474 (2019) 262--268.

\bibitem{lee2020open}
S.~B. Lee, K.~Mitra, H.~D. Pratt~III, T.~M. Anderson, V.~Ramadesigan, B.~R.
  Chalamala, V.~R. Subramanian, Open data, models, and codes for vanadium redox
  batch cell systems: a systems approach using zero-dimensional models, Journal
  of Electrochemical Energy Conversion and Storage 17~(1) (2020) 011008.

\bibitem{vynnycky2011analysis}
M.~Vynnycky, Analysis of a model for the operation of a vanadium redox battery,
  Energy 36~(4) (2011) 2242--2256.

\bibitem{chen2014enhancement}
C.~L. Chen, H.~K. Yeoh, M.~H. Chakrabarti, An enhancement to vynnycky's model
  for the all-vanadium redox flow battery, Electrochimica Acta 120 (2014)
  167--179.

\bibitem{sharma2014quasi}
A.~Sharma, M.~Vynnycky, C.~Ling, E.~Birgersson, M.~Han, The quasi-steady state
  of all-vanadium redox flow batteries: A scale analysis, Electrochimica Acta
  147 (2014) 657--662.

\bibitem{al2009non}
H.~Al-Fetlawi, A.~Shah, F.~Walsh, Non-isothermal modelling of the all-vanadium
  redox flow battery, Electrochimica Acta 55~(1) (2009) 78--89.

\bibitem{you2009simple}
D.~You, H.~Zhang, J.~Chen, A simple model for the vanadium redox battery,
  Electrochimica Acta 54~(27) (2009) 6827--6836.

\bibitem{shah2010dynamic}
A.~Shah, H.~Al-Fetlawi, F.~Walsh, Dynamic modelling of hydrogen evolution
  effects in the all-vanadium redox flow battery, Electrochimica Acta 55~(3)
  (2010) 1125--1139.

\bibitem{knehr2012transient}
K.~Knehr, E.~Agar, C.~Dennison, A.~Kalidindi, E.~Kumbur, A transient vanadium
  flow battery model incorporating vanadium crossover and water transport
  through the membrane, Journal of The Electrochemical Society 159~(9) (2012)
  A1446.

\bibitem{choi2020multiple}
Y.~Y. Choi, S.~Kim, S.~Kim, J.-I. Choi, Multiple parameter identification using
  genetic algorithm in vanadium redox flow batteries, Journal of Power Sources
  450 (2020) 227684.

\bibitem{ma2011three}
X.~Ma, H.~Zhang, F.~Xing, A three-dimensional model for negative half cell of
  the vanadium redox flow battery, Electrochimica Acta 58 (2011) 238--246.

\bibitem{xu2013numerical}
Q.~Xu, T.~Zhao, P.~Leung, Numerical investigations of flow field designs for
  vanadium redox flow batteries, Applied energy 105 (2013) 47--56.

\bibitem{fu2023three}
Y.~Fu, J.~Bao, C.~Zeng, Y.~Chen, Z.~Xu, S.~Kim, W.~Wang, A three-dimensional
  pore-scale model for redox flow battery electrode design analysis, Journal of
  Power Sources 556 (2023) 232329.

\bibitem{zheng2014three}
Q.~Zheng, H.~Zhang, F.~Xing, X.~Ma, X.~Li, G.~Ning, A three-dimensional model
  for thermal analysis in a vanadium flow battery, Applied energy 113 (2014)
  1675--1685.

\bibitem{yin2014coupled}
C.~Yin, Y.~Gao, S.~Guo, H.~Tang, A coupled three dimensional model of vanadium
  redox flow battery for flow field designs, Energy 74 (2014) 886--895.

\bibitem{oh2015three}
K.~Oh, H.~Yoo, J.~Ko, S.~Won, H.~Ju, Three-dimensional, transient,
  nonisothermal model of all-vanadium redox flow batteries, Energy 81 (2015)
  3--14.

\bibitem{yin2015numerical}
C.~Yin, S.~Guo, H.~Fang, J.~Liu, Y.~Li, H.~Tang, Numerical and experimental
  studies of stack shunt current for vanadium redox flow battery, Applied
  Energy 151 (2015) 237--248.

\bibitem{messaggi2018analysis}
M.~Messaggi, P.~Canzi, R.~Mereu, A.~Baricci, F.~Inzoli, A.~Casalegno, M.~Zago,
  Analysis of flow field design on vanadium redox flow battery performance:
  Development of 3d computational fluid dynamic model and experimental
  validation, Applied energy 228 (2018) 1057--1070.

\bibitem{goodfellow2016deep}
I.~Goodfellow, Y.~Bengio, A.~Courville, Deep learning, MIT press, 2016.

\bibitem{artrith2019machine}
N.~Artrith, Machine learning for the modeling of interfaces in energy storage
  and conversion materials, Journal of Physics: Energy 1~(3) (2019) 032002.

\bibitem{gao2021machine}
T.~Gao, W.~Lu, Machine learning toward advanced energy storage devices and
  systems, Iscience 24~(1) (2021) 101936.

\bibitem{chen2020machine}
A.~Chen, X.~Zhang, Z.~Zhou, Machine learning: accelerating materials
  development for energy storage and conversion, InfoMat 2~(3) (2020) 553--576.

\bibitem{wan2021coupled}
S.~Wan, X.~Liang, H.~Jiang, J.~Sun, N.~Djilali, T.~Zhao, A coupled machine
  learning and genetic algorithm approach to the design of porous electrodes
  for redox flow batteries, Applied Energy 298 (2021) 117177.

\bibitem{li2020cost}
T.~Li, F.~Xing, T.~Liu, J.~Sun, D.~Shi, H.~Zhang, X.~Li, Cost, performance
  prediction and optimization of a vanadium flow battery by machine-learning,
  Energy \& Environmental Science 13~(11) (2020) 4353--4361.

\bibitem{bao2020machine}
J.~Bao, V.~Murugesan, C.~J. Kamp, Y.~Shao, L.~Yan, W.~Wang, Machine learning
  coupled multi-scale modeling for redox flow batteries, Advanced Theory and
  Simulations 3~(2) (2020) 1900167.

\bibitem{pang2023physics}
H.~Pang, L.~Wu, J.~Liu, X.~Liu, K.~Liu, Physics-informed neural network
  approach for heat generation rate estimation of lithium-ion battery under
  various driving conditions, Journal of Energy Chemistry 78 (2023) 1--12.

\bibitem{dakshinamoorthy2023estimating}
H.~Dakshinamoorthy, V.~L. Srinivas, Estimating battery temperature in dynamic
  driving conditions using physics informed neural networks, in: 2023 IEEE IAS
  Global Conference on Emerging Technologies (GlobConET), IEEE, 2023, pp. 1--6.

\bibitem{deng2023physics}
H.-P. Deng, Y.-B. He, B.-C. Wang, H.-X. Li, Physics-dominated neural network
  for spatiotemporal modeling of battery thermal process, IEEE Transactions on
  Industrial Informatics (2023).

\bibitem{cho2022physics}
G.~Cho, M.~Wang, Y.~Kim, J.~Kwon, W.~Su, A physics-informed machine learning
  approach for estimating lithium-ion battery temperature, IEEE Access 10
  (2022) 88117--88126.

\bibitem{wen2023fusing}
P.~Wen, Z.-S. Ye, Y.~Li, S.~Chen, S.~Zhao, Fusing models for prognostics and
  health management of lithium-ion batteries based on physics-informed neural
  networks, arXiv preprint arXiv:2301.00776 (2023).

\bibitem{sun2023adaptive}
B.~Sun, J.~Pan, Z.~Wu, Q.~Xia, Z.~Wang, Y.~Ren, D.~Yang, X.~Guo, Q.~Feng,
  Adaptive evolution enhanced physics-informed neural networks for time-variant
  health prognosis of lithium-ion batteries, Journal of Power Sources 556
  (2023) 232432.

\bibitem{huang2023minn}
Y.~Huang, C.~Zou, Y.~Li, T.~Wik, Minn: Learning the dynamics of
  differential-algebraic equations and application to battery modeling, arXiv
  preprint arXiv:2304.14422 (2023).

\bibitem{wang2022physics}
Y.~Wang, X.~Han, D.~Guo, L.~Lu, Y.~Chen, M.~Ouyang, Physics-informed recurrent
  neural network with fractional-order gradients for state-of-charge estimation
  of lithium-ion battery, IEEE Journal of Radio Frequency Identification 6
  (2022) 968--971.

\bibitem{he2022physics}
Q.~He, P.~Stinis, A.~M. Tartakovsky, Physics-constrained deep neural network
  method for estimating parameters in a redox flow battery, Journal of Power
  Sources 528 (2022) 231147.

\bibitem{he2022enhanced}
Q.~He, Y.~Fu, P.~Stinis, A.~Tartakovsky, Enhanced physics-constrained deep
  neural networks for modeling vanadium redox flow battery, Journal of Power
  Sources 542 (2022) 231807.

\bibitem{howard2022physics}
A.~A. Howard, T.~Yu, W.~Wang, A.~M. Tartakovsky, Physics-informed cokriging
  model of a redox flow battery, Journal of Power Sources 542 (2022) 231668.

\bibitem{bates2015modeling}
A.~Bates, S.~Mukherjee, N.~Schuppert, B.~Son, J.~G. Kim, S.~Park, Modeling and
  simulation of 2d lithium-ion solid state battery, International Journal of
  Energy Research 39~(11) (2015) 1505--1518.

\bibitem{santhanagopalan2006review}
S.~Santhanagopalan, Q.~Guo, P.~Ramadass, R.~E. White, Review of models for
  predicting the cycling performance of lithium ion batteries, Journal of power
  sources 156~(2) (2006) 620--628.

\bibitem{abadi2016tensorflow}
M.~Abadi, P.~Barham, J.~Chen, Z.~Chen, A.~Davis, J.~Dean, M.~Devin,
  S.~Ghemawat, G.~Irving, M.~Isard, et~al., Tensorflow: a system for
  large-scale machine learning., in: Osdi, Vol.~16, Savannah, GA, USA, 2016,
  pp. 265--283.

\bibitem{jokar2016review}
A.~Jokar, B.~Rajabloo, M.~D{\'e}silets, M.~Lacroix, Review of simplified
  pseudo-two-dimensional models of lithium-ion batteries, Journal of Power
  Sources 327 (2016) 44--55.

\bibitem{fu2023understanding}
Y.~Fu, R.~K. Singh, S.~Feng, J.~Liu, J.~Xiao, J.~Bao, Z.~Xu, D.~Lu,
  Understanding of low-porosity sulfur electrode for high-energy
  lithium--sulfur batteries, Advanced Energy Materials 13~(13) (2023) 2203386.

\bibitem{kumaresan2008mathematical}
K.~Kumaresan, Y.~Mikhaylik, R.~E. White, A mathematical model for a
  lithium--sulfur cell, Journal of the electrochemical society 155~(8) (2008)
  A576.

\bibitem{danilov2010modeling}
D.~Danilov, R.~Niessen, P.~Notten, Modeling all-solid-state li-ion batteries,
  Journal of the Electrochemical Society 158~(3) (2010) A215.

\bibitem{qiu20123}
G.~Qiu, A.~S. Joshi, C.~Dennison, K.~Knehr, E.~Kumbur, Y.~Sun, 3-d pore-scale
  resolved model for coupled species/charge/fluid transport in a vanadium redox
  flow battery, Electrochimica Acta 64 (2012) 46--64.

\bibitem{wang2021understanding}
S.~Wang, Y.~Teng, P.~Perdikaris, Understanding and mitigating gradient flow
  pathologies in physics-informed neural networks, SIAM Journal on Scientific
  Computing 43~(5) (2021) A3055--A3081.

\bibitem{loffe2014accelerating}
S.~Loffe, C.~Normalization, Accelerating deep network training by reducing
  internal covariate shift, arXiv (2014).

\bibitem{mcclenny2020self}
L.~McClenny, U.~Braga-Neto, Self-adaptive physics-informed neural networks
  using a soft attention mechanism, arXiv preprint arXiv:2009.04544 (2020).

\bibitem{ramachandran2017searching}
P.~Ramachandran, B.~Zoph, Q.~V. Le, Searching for activation functions, arXiv
  preprint arXiv:1710.05941 (2017).

\bibitem{glorot2010understanding}
X.~Glorot, Y.~Bengio, Understanding the difficulty of training deep feedforward
  neural networks, in: Proceedings of the thirteenth international conference
  on artificial intelligence and statistics, JMLR Workshop and Conference
  Proceedings, 2010, pp. 249--256.

\bibitem{kingma2014adam}
D.~P. Kingma, J.~Ba, Adam: A method for stochastic optimization, arXiv preprint
  arXiv:1412.6980 (2014).

\bibitem{liu1989limited}
D.~C. Liu, J.~Nocedal, On the limited memory bfgs method for large scale
  optimization, Mathematical programming 45~(1-3) (1989) 503--528.

\bibitem{paszke2019pytorch}
A.~Paszke, S.~Gross, F.~Massa, A.~Lerer, J.~Bradbury, G.~Chanan, T.~Killeen,
  Z.~Lin, N.~Gimelshein, L.~Antiga, et~al., Pytorch: An imperative style,
  high-performance deep learning library, Advances in neural information
  processing systems 32 (2019).

\bibitem{dickinson2014comsol}
E.~J. Dickinson, H.~Ekstr{\"o}m, E.~Fontes, Comsol
  multiphysics{\textregistered}: Finite element software for electrochemical
  analysis. a mini-review, Electrochemistry communications 40 (2014) 71--74.

\end{thebibliography}

\end{document}